\documentclass[runningheads]{llncs}
\usepackage{graphicx}
\usepackage{times}
\usepackage{epsfig}
\usepackage{graphicx}
\usepackage{amsmath}
\usepackage{amssymb}
\usepackage{enumitem}
\usepackage[belowskip=-12pt,aboveskip=-6pt]{caption}
\usepackage[pagebackref=true,breaklinks=true,colorlinks,bookmarks=false]{hyperref}
\usepackage{makecell}
\usepackage{array}
\usepackage{cite}
\usepackage[toc, page]{appendix}
\newcolumntype{?}{!{\vrule width 1pt}}
\begin{document}
\title{Targeted Gradient Descent: A Novel Method for Convolutional Neural Networks Fine-tuning and Online-learning}
\titlerunning{TGD: A Novel Method for Fine-tuning ConvNets}
%
\author{Junyu Chen\inst{1, 2} \and
Evren Asma\inst{3} \and
Chung Chan\inst{3}}
%
\authorrunning{MICCAI 2021 submission - J. Chen et al.}
%
\institute{Department of Radiology and Radiological Science, Johns Hopkins University, MD, USA \and
Department of Electrical and Computer Engineering, Johns Hopkins University, MD, USA \and
Canon Medical Research USA, INC., IL, USA\\
\email{jchen245@jhmi.edu, \{easma,cchan\}@mru.medical.canon}}
\maketitle              
\begin{abstract}
A convolutional neural network (ConvNet) is usually trained and then tested using images drawn from the same distribution. To generalize a ConvNet to various tasks often requires a complete training dataset that consists of images drawn from different tasks. In most scenarios, it is nearly impossible to collect every possible representative dataset as a priori. The new data may only become available after the ConvNet is deployed in clinical practice. ConvNet, however, may generate artifacts on out-of-distribution testing samples. In this study, we present Targeted Gradient Descent (TGD), a novel fine-tuning method that can extend a pre-trained network to a new task without revisiting data from the previous task while preserving the knowledge acquired from previous training. To a further extent, the proposed method also enables online learning of patient-specific data. The method is built on the idea of reusing a pre-trained ConvNet’s redundant kernels to learn new knowledge. We compare the performance of TGD to several commonly used training approaches on the task of Positron emission tomography (PET) image denoising. Results from clinical images show that TGD generated results on par with training-from-scratch while significantly reducing data preparation and network training time. More importantly, it enables online learning on the testing study to enhance the network’s generalization capability in real-world applications.

\end{abstract}

\section{Introduction}

A Convolutional neural network (ConvNet) is usually trained and tested on datasets where images were sampled from the same distribution. However, if training and testing images are drawn from different distributions, ConvNets would suffer from performance degradation. This is a commonly observed scenario in medical imaging applications due to variations among patients, image acquisition, and reconstruction protocols \cite{Chen2018, Kamnitsas2017, Ghafoorian2017}. For example, when applying denoising ConvNets on unseen features, it may cause artifacts in the denoised images as demonstrated in both Ultrasound \cite{laves2020uncertainty} and Positron Emission Tomography (PET) applications \cite{chan2020est}. To generalize a trained ConvNet to different image distributions, one has to include images sampled from the new distribution (task) in the training dataset and retrain the ConvNet. However, in medical imaging, generating labeled datasets is often tedious, time-consuming, and expensive. In most scenarios, it is nearly impossible to collect every possible representative dataset as a priori. \textit{In denoising applications, the new data without high quality label may only become available after the ConvNet is deployed (as the high quality label usually requires extra radiation dose or prolonged scan duration).} Moreover, in product development, there is often a need to improve the imaging protocols (i.e., scan or reconstruction protocol) during a product development phase. The change of the image properties, i.e., local pixel correlations, would require regenerating all the training datasets with the updated protocol followed by retraining the denoising network. This recurring process is inevitably time and resource-consuming. Thus, it is more desirable to develop methods that can adapt to various image distributions with minimum need for additional training data and training time. Ultimately, the goal is to develop an online learning algorithm that can quickly retrain and adapt a pre-trained ConvNet to each testing dataset.

Fine-tuning is a promising approach to avoid training a ConvNet from scratch. During fine-tuning, a pre-trained network, usually trained using a large number of datasets from a different application, is used to continue the backpropagation on a smaller dataset from a new task \cite{amiri2019fine, gong2018pet}. However, fine-tuning the network on a new task does not guarantee retaining the useful knowledge acquired from the previous training. If the number of training dataset from the new task is much less than that used in the old task, the fine-tuned network will overfit to the datasets in the new task with degraded generalization capability \cite{hinton2015distilling}, which may not be suitable for the applications in which both tasks are of interest during testing. Another approach is using joint training (e.g., \cite{caruana1997multitask, wu2018memory}) or incremental learning (e.g., \cite{castro2018end, xiao2014error, tasar2019incremental, rusu2016progressive}). They try to adapt a pre-trained network to new tasks while preserving the network’s original capabilities. Joint training requires revisiting data from previous tasks while learning the new task \cite{li2017learning, caruana1997multitask, wu2018memory} or modifying the network’s architecture \cite{rusu2016progressive}. Continual learning is used to continuously adapt a ConvNet to a constantly arriving data stream, enabling the ConvNet to learn new tasks incrementally and without forgetting the ones already learned. McClure et al. proposed a continual learning method that consolidates the weights of separate neural networks \cite{mcclure2018distributed}. The method necessitates that the networks to be trained on the complete datasets. However, obtaining such data may not always be possible. Another example is Elastic Weight Consolidation (EWC) \cite{kirkpatrick2017overcoming, baweja2018towards}, which uses Fisher Information Matrix to regularize the penalty term when fine-tuning an existing network using new datasets. Although this method does not require the old training dataset, it might be difficult to fine-tune the hyper-parameter to balance the strength of the weight regularizer and the loss of the new task, especially when only a single testing dataset without label is available.
\begin{figure*}[b]
\begin{center}
\includegraphics[width=1\textwidth]{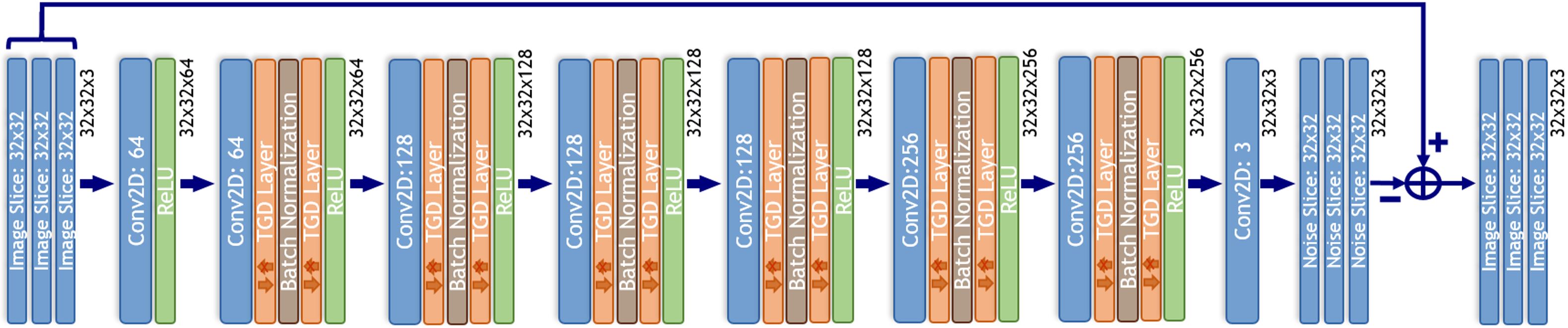}
\end{center}
   \caption{The architecture of an eight-layer denoising residual network with proposed TGD layers (orange blocks).}
\label{fig:arch}
\end{figure*}

Instead of blindly fine-tuning all the kernels in the specific layers or retraining the entire network with a mixture of new and old labels, it might be more sensible to precisely retrain the “meaningless” kernels to make them adapt to the new tasks while the ``useful” kernels are preserved so they can retain the knowledge acquired from the prior training with a larger training dataset (a wider coverage of data distribution). This work proposes a novel fine-tuning method, the Targeted Gradient Descent (TGD) layer, that can be inserted into any ConvNet architecture. The novel contributions of the proposed method are 2-fold: 1. TGD can extend a pre-trained network to a new task without revisiting data from the previous task while preserving the knowledge acquired from previous training; 2. It enables online learning that adapts a pre-trained network to each testing dataset to avoid generating artifacts on unseen features. We demonstrate the proposed method’s effectiveness in denoising tasks for PET images.

\section{Methodology}

In this study, the pre-trained PET denoising ConvNet was built on the basis of the denoising convolutional neural network (DnCNN) \cite{zhang2017beyond}. The architecture of the DnCNN is the same as in Fig. \ref{fig:arch} but \textit{without} the TGD layers. It is a 2.5-dimensional network that takes three consecutive 2D image slices as its input. The network consists of eight 3$\times$3 convolutional layers and a single residual layer at the end of the network. Each convolutional layer is followed by a batch normalization and a rectified linear unit (ReLU), except for the first and the last convolutional layers. The first convolution layer is followed by a ReLU, whereas the last convolution layer is not followed by any activation. 

To update the specific kernels in the fine-tuning training, we first need to determine the information richness in the feature maps. The corresponding network kernels can then be identified and updated in the retraining stage to generate new feature maps, such that if the kernels can produce meaningful features, which are identified as ``useful" kernels, while the kernels producing ``meaningless" features are identified as ``meaningless" kernels. However, it is hard to determine a feature map’s information richness based solely on some particular input images because different input images may activate different feature maps. Here we used Kernel Sparsity and Entropy (KSE) metric proposed by Li et al. \cite{li2019exploiting}. The KSE quantifies the sparsity and information richness in a kernel to evaluate a feature map’s importance to the network. The KSE contains two parts: the kernel sparsity, $s_c$, and the kernel entropy, $e_c$, and they are briefly described here. We refer readers to \cite{li2019exploiting} for details. The \textbf{kernel sparsity} for the $c^{th}$ input feature map is defined as: 

\begin{equation}
    s_c = \sum^{N}_{n=1}\vert W_{n,c}\vert,
\end{equation}
\noindent where $N$ denotes the total number output feature maps, $W_{n,c}$ denotes the 2D kernels, $n$ and $c$ are, respectively, the indices of the output and input feature maps. The \textbf{kernel entropy} is calculated as the entropy of the density metrics (i.e., $dm(\cdot)$):
\begin{equation}
    e_c = -\sum^{N}_{i=1}\frac{dm(W_{i,c})}{\sum_{i=1}^{N}dm(W_{i,c})}\log_2\frac{dm(W_{i,c})}{\sum_{i=1}^{N}dm(W_{i,c})},
\end{equation}
where $dm(W_{i,c})=\sum_jA_{c_{ij}}$, and $A_{c_{ij}}$ is a nearest neighbor distance matrix for the $c^{th}$ convolutional kernel \cite{li2019exploiting}. A small $e_c$ indicates diverse convolution kernels. Thus, the corresponding input feature map provides more information to the ConvNet. KSE is then defined as:
\begin{equation}
    KSE = \sqrt{\frac{s_c}{1+\alpha e_c}},
\end{equation}
where $KSE$, $s_c$, and $e_c$ are normalized into [0, 1], and $\alpha$ is a parameter for controlling weight between $s_c$ and $e_c$, which is set to 1 according to \cite{li2019exploiting}.

\begin{figure}[t]
\begin{center}
\includegraphics[width=1\textwidth]{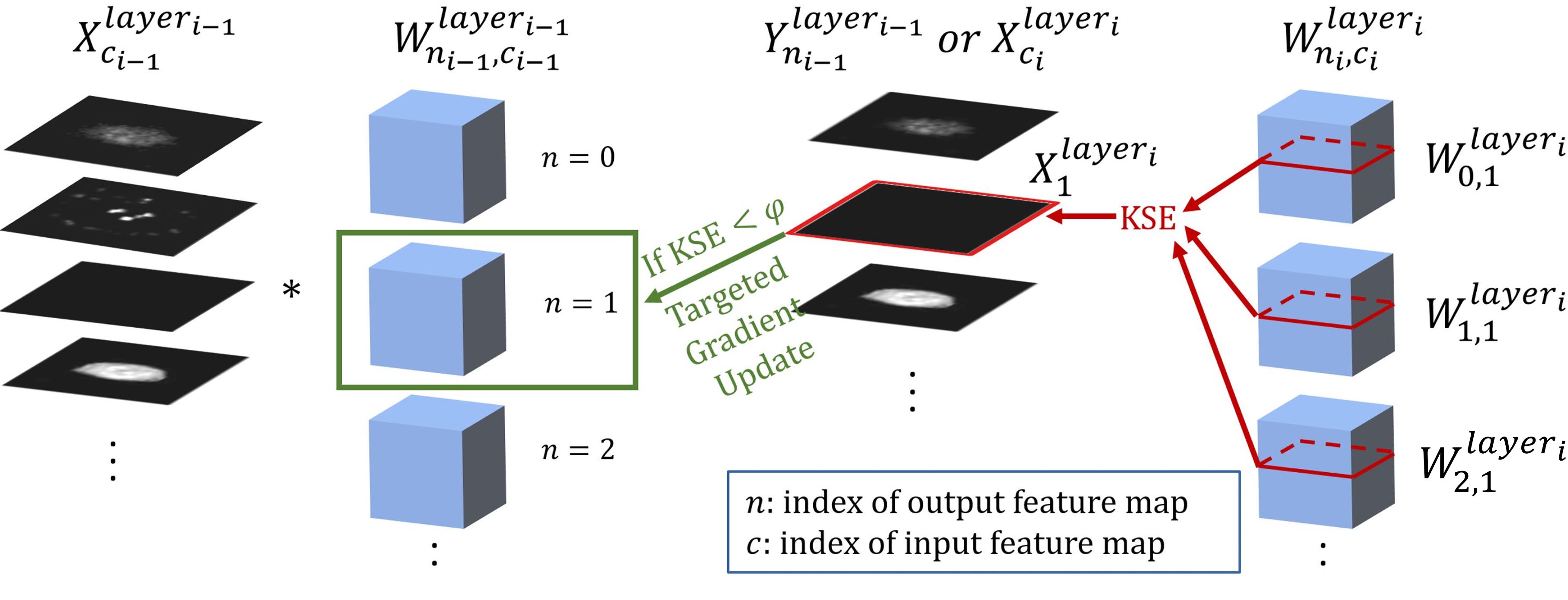}
\end{center}
   \caption{The framework of identifying kernels that generate redundant feature maps. The kernel weights in layer $i$ (i.e., $W_{n_i,c_i}^{\text{layer}_{i}}$) were used to calculate KSE scores for the input feature maps in layer $i$ (i.e., $X_{c_i}^{\text{layer}_{i}}$), then the kernels in layer $i-1$ (e.g., the green box: $W_{1,c_{i-1}}^{\text{layer}_{i-1}}$) that generated the input feature maps in layer $i$ (i.e., $X_{c_i}^{layer_i}$) were identified and would be retrained in the proposed TGD method.}
\label{fig:simple_conv}
\end{figure}

\subsection{Targeted Gradient Descent Layer}
\label{sec:TGD}
The KSE score indicates the meaningfulness of feature maps to the ConvNet. Our goal is to retrain the kernels that generate redundant feature maps and keep the “useful” kernels unchanged. In this paper, we denote $X$ and $Y$, respectively, to be the input and output feature maps of a convolutional layer. As illustrated in Fig. \ref{fig:simple_conv}, we first calculate KSE for the input feature maps of layer $i$ using the corresponding kernel weights from the $i^{th}$ convolutional layer. The feature maps with KSE scores below a certain user-defined threshold, $\phi$, are marked as meaningless. We then identify and record the indices of the convolution kernels that generate the “meaningless” feature maps from the $(i-1)^{th}$ layer. The indices were used for creating a binary mask, $M$:
\[
    M_n= 
\begin{cases}
    \mathbf{1},& \text{if } KSE(Y_n) < \phi\\
    \mathbf{0},              & \text{if } KSE(Y_n) \geq \phi
\end{cases},
\]
\noindent where $\phi$ is the user-defined KSE threshold. $M_{n}$ zeros out the gradients for the "useful" kernels (i.e.,  $KSE(Y_n)\geq \phi$), so that these kernels will not be modified during retraining. The back-propagation formula is then adjusted to incorporate $M_{n}$ as:
\begin{equation}
\begin{split}
W^{(t+1)}_{n,c} = W^{(t)}_{n} &- \eta\frac{\partial\mathcal{L}}{\partial Y^{(t)}_n}M_{n}X^{(t)}_c - \frac{\partial\mathcal{R}(W^{(t)}_{n})}{\partial Y^{(t)}_n}M_{n}X^{(t)}_c
\end{split},
\end{equation}
\noindent where $\mathcal{L}$ and $\mathcal{R}$ denote, respectively, the loss function and weight regularization. We embedded the gradient zeroing process (i.e., $\frac{\partial\mathcal{L}}{\partial Y^{(t)}_n}M_{n}$) into a novel layer, named Targeted Gradient Descent layer (the orange blocks in Fig. \ref{fig:arch}). Notice that the batch normalization layers contain trainable weights as well (i.e., $\gamma$, $\beta$, and $\sigma^2$), where their gradients, $\frac{\partial\mathcal{L}}{\partial\gamma^{(t)}}$, $\frac{\partial\mathcal{L}}{\partial\beta^{(t)}}$, and $\frac{\partial\mathcal{L}}{\partial\sigma^{2^{(t)}}}$ at iteration $t$ can also be expressed as a function of $\frac{\partial\mathcal{L}}{\partial Y^{(t)}_n}$. As a result, the TGD layer was inserted after each convolutional layer as well as each batch normalization layer. Note that \textit{the TGD layers are disabled during forward pass which means all kernels are activated.} During back-propagation, the TGD layers are activated and only the targeted kernels are updated. The final architecture of the TGD-net is shown in Fig. \ref{fig:arch}.

The TGD-net adapted from a first task to the second task can further be adapted to a third task. The same TGD retraining process can be applied to the retrained TGD-net again, i.e., calculate KSE scores for the feature maps in the TGD-net, form new gradient masks, $M_n$, for the TGD layers, and then retrain the TGD-net with images from a third task. We name this recurring process $\text{TGD}^n$, where n represents the number of TGD retraining processes applied to a single network. In this work, we evaluated the cases in which $n=1 \& 2$ (i.e., TGD and $\text{TGD}^2$).

\begin{figure}[t]
\begin{center}
\includegraphics[width=0.6\textwidth]{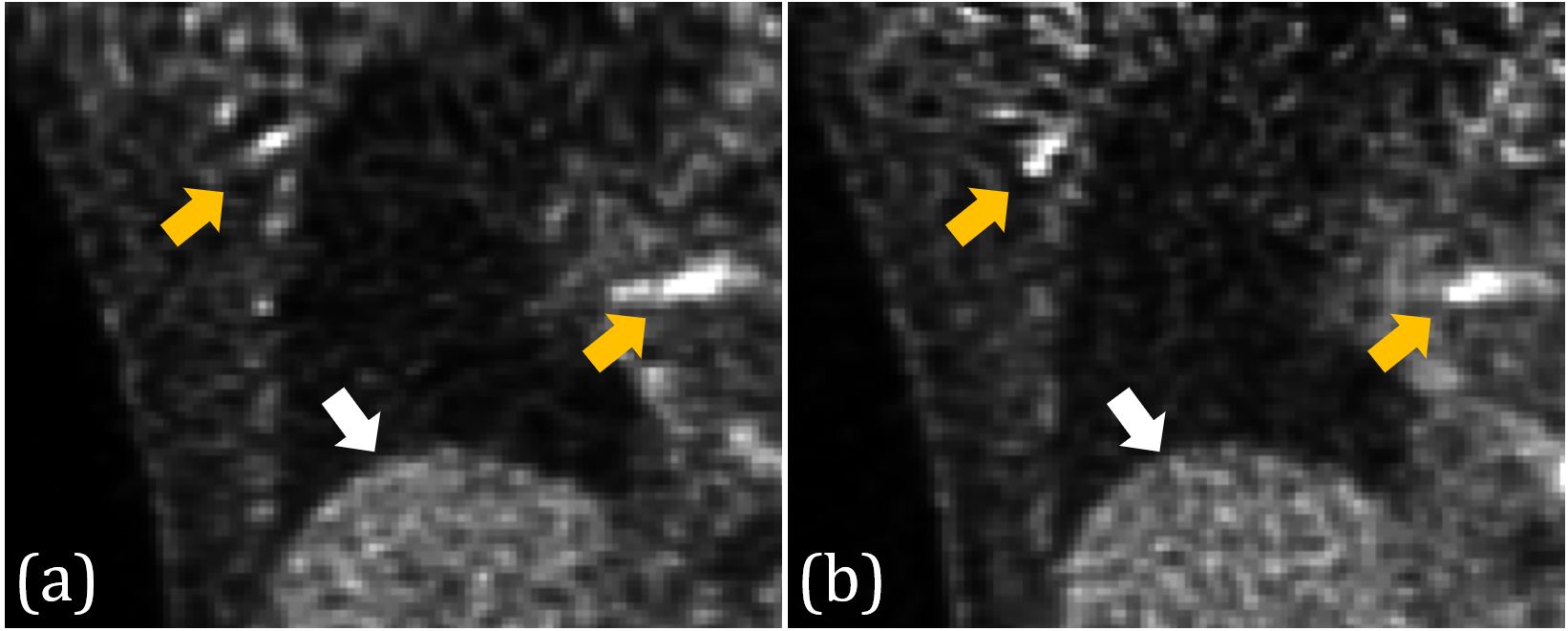}
\end{center}
   \caption{The PET images of the right lung of a human subject reconstructed with (a) v1 reconstruction and (b) v2 reconstruction methods, respectively. The yellow arrows denote the structures that become elongated in v1 images due to the sub-optimal resolution modeling, which are corrected in the v2 reconstruction. The optimized resolution modeling also produced the liver noise texture (denoted by the white arrows) with a finer grain size (which represents a small correlation between neighboring pixels) in the (b) than (a).}
\label{fig:v1_v2}
\end{figure}

\section{TGD in PET Denoising}
We demonstrate the proposed TGD method on the task of PET image denoising in two applications. We first use TGD to fine-tune an existing denoising ConvNet to make it adapt to a new reconstruction protocol using substantially fewer training studies. We then further use TGD in an \textit{online-learning} approach to avoid the ConvNet generating artifacts (hallucinations) on unseen features during testing. 

\subsection{Fine-tuning}
The pre-trained network was trained using FDG PET images acquired on a commercial SiPM PET/CT scanner reconstructed from a prior version of the ordered subset expectation maximization (OSEM) algorithm. For simplicity, we denote these images as the v1 images and the denoising ConvNet trained using these images as the v1 network. We denote the PET images reconstructed by an updated OSEM algorithm as the v2 images and the corresponding denoising ConvNet as the v2 network. The system resolution modeling and scatter estimation in v2 reconstruction were optimized over the v1 reconstruction. Therefore, the noise texture in v2 images is finer, indicating a smaller correlation among neighboring pixels as shown in Fig. \ref{fig:v1_v2}. The application of the v1 network on v2 images produced over-smoothed results and suppressed activity in small lesions, which could potentially lead to misdiagnosis.

Conventionally, whenever the reconstruction algorithm is updated, the entire training datasets have to be re-reconstructed, and the denoising network has to be retrained using the updated images for optimal performance, followed by qualitative and quantitative assessments on a cohort of testing studies. This process is extremely tedious and time-consuming.

The v1 network was trained using 20 $\times$ v1 whole-body FDG-PET human studies with a mixture of low ($<23$) and high BMI ($>28$).  These studies were acquired for 10-min/bed, which were used as the target images. We uniformly subsampled the list mode data into 6 noise levels as 30, 45, 60, 90, 120, 180 sec/bed as the noisy inputs for noise adaptive training \cite{chan2018noise}. All these studies consist of 30,720 training slices in total. This v1 network was adapted using the TGD method to denoising v2 PET images. During the TGD’s retraining stage, we used only 7 training datasets that consist of PET scans from patients with low BMI ($<23$). However, the retrained network retained the knowledge on how to denoise PET scans of high BMI patients learned from the previous task (images of high BMI subjects are commonly substantially noisier than those of low BMI subjects). It is important to emphasize that the amount of v1 images used in v1 network training was significantly more than the amount of v2 images used in TGD fine-tuning. Based on this fact, we kept the weights of the noise classifier layer (i.e., the last convolutional layer) in the TGD-net unchanged during the retraining, thus avoiding the last layer from being biased by the v2 image data.

\subsection{Online-learning}
In the second experiment, we showed that TGD enables online-learning that further optimize the network’s performance on each testing study and prevents artifacts (hallucination) from occurring on out-of-distribution features. This is achieved by using TGD with Noise-2-Noise (N2N) training scheme \cite{lehtinen2018noise2noise, chan2019noise}. Specifically, we rebinned a testing study list-mode data acquired with 120-sec into 2 noise realizations with equal count levels (60-sec). We used TGD to fine-tune the denoising network by using noise samples 1 and 2 as the inputs and noise samples 2 and 1 as the targets. We denote the online-learning network as TGD$_{\text{N2N}}$-net. To a greater extent, this procedure was also applied to the TGD-net from the first experiment (the network was TGD fine-tuned twice), and we denote the resulting network as TGD$^2_{\text{N2N}}$-net for convenience.

\begin{figure}[t]
\begin{center}
\includegraphics[width=1\textwidth]{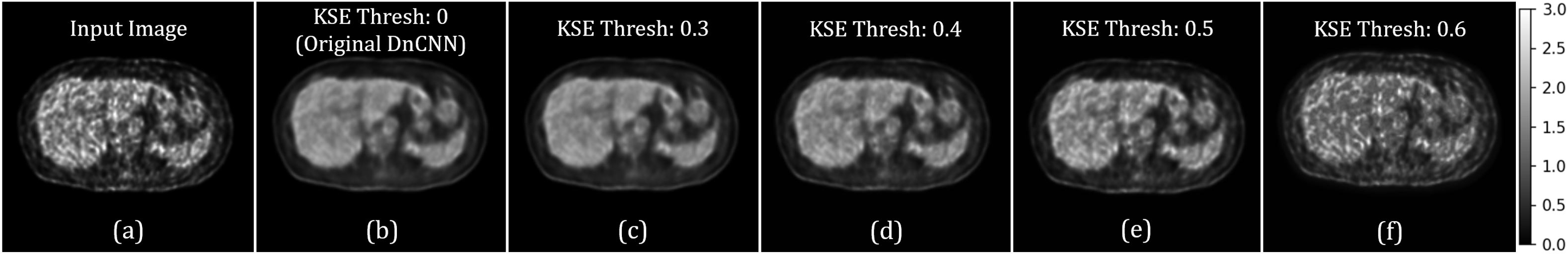}
\end{center}
   \caption{The network's outputs produced by varying KSE threshold.}
\label{fig:KSE_thresh}
\end{figure}

\section{Experiments}
The optimal KSE threshold was first studied. During the prediction, we dropped the kernels (i.e., setting the weights to zeros) identified as “meaningless” by the KSE threshold to examine whether these kernels indeed contributed less to the network. As shown in Fig. \ref{fig:KSE_thresh}, where (a) shows an example slice of a PET scan, and (b) shows the denoised PET image from the v1 DnCNN (this can be interpreted as having a KSE threshold = 0, because no kernel was dropped). We then arbitrarily tested four thresholds: 0.3, 0.4, 0.5, and 0.6. The larger the thresholds, the more the kernels were dropped. The percentage of the parameters that were dropped using the four thresholds are, respectively, 51.4\%, 51.6\%. 53.0\%, and 67.3\%, and the corresponding denoised results are shown in (c), (d), (e), and (f) of Fig. \ref{fig:KSE_thresh}, respectively. The result from $\phi=0.3$ is almost identical to the original DnCNN’s result. Whereas, when  $\phi>0.4$, some severe artifacts begin to occur in the resulting images. Therefore, the KSE threshold, $\phi$, was set to be 0.3 in this work.

TGD-net was compared to several baseline methods, including: 1. v1-net: A DnCNN trained using $20\times$ v1 PET images; 2. v2-net: A DnCNN trained using the same 20 studies but reconstructed with v2 algorithm; 3. FT-net: Fine-tuning the last three convolutional blocks \cite{gong2018pet} of v1-net using only $7\times$ v2 images; 4. TGD-net: v1-net fine-tuned using the TGD layers with $7\times$ v2 images (same studies as used in the FT-net). All networks were trained with 500 epochs.

The proposed $\text{TGD}_{\text{N2N}}^2$-net$^{\phi=0.3,0.4}$ and $\text{TGD$_{\text{N2N}}$-net}^{\phi=0.4}$ were built based on the previous TGD-net and v2-net, respectively. These networks were retrained using two noise realizations from a single study (i.e., N2N training). They were compared to: 1. v2-net: Same as above. 2. $\text{TGD-net}^{\phi=0.3}$: The TGD-net obtained from the previous task. The TGD$_{\text{N2N}}$ models were trained with 150 epochs. 

All these methods were compared in terms of denoising on 3 FDG patient studies (2 are shown in the results, and 1 is shown in the supplementary materials) reconstructed with v2 algorithm (v2 images). One of the studies was acquired with 600-sec/bed with a simulated tumor that was inserted in the liver. We rebinned the list-mode study into 10 $\times$ 60-sec/bed image i.i.d noise realizations to assess the ensemble bias on the tumor and liver coefficient of variation (CoV) by using the 600-sec/bed image as the ground truth. We then further evaluated the methods on a second 60-sec/bed study. 
%
\begin{table}[t]
\small
\begin{center}
\begin{tabular}{c|c|c|c?c}
\hline
& $\text{v1-net}$  & $\text{v2-net}$ & $\text{FT-net}$ &$\text{TGD-net}^{0.3}$\\
\hline
Lesion Bias (\%) & -6.30 &  -4.07 &  -4.71& \textbf{-3.77}\\
\hline
Liver CoV (\%) & 6.02 & 8.56 & 7.87& 6.46\\
\hline
\end{tabular}
\end{center}
\caption{Ensemble bias and CoV comparison results of the different training methods on the PET scan of the low-BMI patient. \label{lowBMI_TGD}}
\end{table}
\subsection{Evaluation of TGD on Whole-Body PET Denoising}
\begin{figure*}[t]
\begin{center}
\includegraphics[width=1\textwidth]{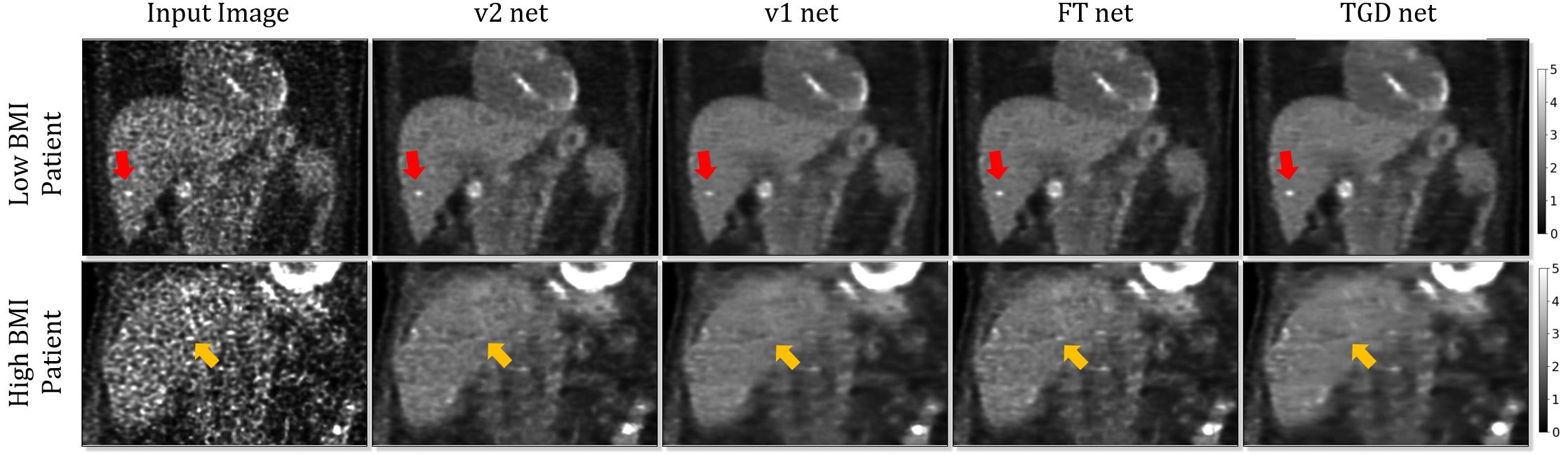}
\end{center}
   \caption{Qualitative comparisons between the proposed TGD method and other methods on denoising two FDG patient studies. The red arrows indicate the synthetic lesion, and the yellow arrows highlight the difference.}
\label{fig:TGD}
\end{figure*}

Fig. \ref{fig:TGD} shows the denoised results of the example cropped slices of the v2 PET images, where the figures in the first column represent the input image. Qualitatively, $\text{v1-net}$ (the third column of Fig. \ref{fig:TGD}) over-smoothed the v2 image that led to piece-wise smoothness in the liver and reduced uptake in the synthetic lesion compared to the results from other methods. In contrast, the result from $\text{v2-net}$ (the second column) exhibited a higher lesion contrast with more natural noise texture (fine grain size) in liver regions. The fine-tuned networks ($\text{FT-net}$) yielded good performances on denoising the low-BMI-patient PET scans (the top figure of the fourth column) with higher lesion contrast. However, the speckle noise (denoted by the yellow arrow) in the high-BMI-patient PET scans was also preserved. The proposed $\text{TGD-net}$ yielded good lesion contrast but also low variations in the liver for both low- and high-BMI patient scans. The quantitative evaluations are shown in Table \ref{lowBMI_TGD}. The best performance is highlighted in bold. For the low-BMI patient study, the proposed method ($\phi=0.3$) achieved the best lesion quantification with a small ensemble bias of -3.77\% while maintaining a low-noise level of 6.45\% in terms of CoV. In addition, fine-tuning a TGD net from the v1-net saved 64\% of computational time compared to the training-from-scratch v2-net.
\begin{figure*}[t]
\begin{center}
\includegraphics[width=0.95\textwidth]{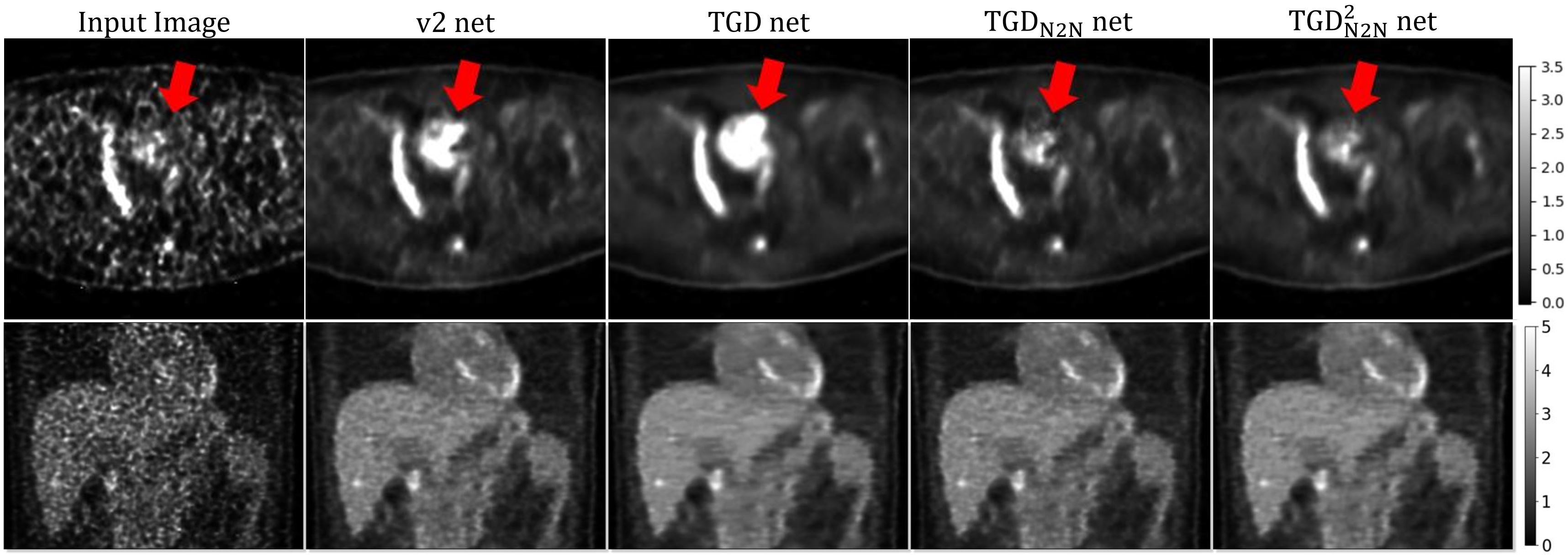}
\end{center}
   \caption{Qualitative comparisons between the proposed TGD$_\text{N2N}$ online learning and other methods. The red arrows indicate the artifactual feature generated by the v2-net and TGD-net around the bladder, which was not included in any training datasets. Both TGD$_\text{N2N}$ and TGD$^2_\text{N2N}$ yielded images which are in high fidelity to the input image on the bladder, while retaining similar denoising performance as v2-net and TGD-net. }
\label{fig:TGD_n2n}

\end{figure*}

Fig. \ref{fig:TGD_n2n} shows the denoised results of the example cropped slices of the v2 PET images. Prior to TGD-N2N training, $\text{v2-net}$ and $\text{TGD-net}$ created artifactual features (hallucination) around the bladder region (denoted by the red arrows). In contrast, the networks fine-tuned using the TGD-N2N online learning scheme did not produce any artifacts, where the bladder's shape is nearly the same as that of the input image. To a greater extend, $\text{TGD}_{\text{N2N}}^2$-net$^{\phi=0.3,0.4}$ and $\text{TGD$_{\text{N2N}}$-net}^{\phi=0.4}$ retained the denoising performances of their base networks (i.e., v2-net and TGD-net). An additional sample patient study with urinary catheter is shown in the Suppl. Material Fig. 1.

\section{Conclusion}

This study introduced Targeted Gradient Descent, a novel incremental learning scheme that effectively reuses the redundant kernels in a pre-trained network. The proposed method can be easily inserted as a layer into an existing network and \textit{does not require revisiting the data from the previous task.} More importantly, it may enable \textit{online learning} on the testing study to enhance the network’s generalization capability in real-world applications. 
%
%
%

\bibliography{egbib}
\clearpage
\section*{Supplemental Materials}
\label{appd}
\subsection{Kernel Sparsity ($\mathbf{s_c}$)}
During the ConvNet backpropagation, the update of the 2D kernels, $W_{n,c}$ (where $n$ and $c$ are, respectively, the indices of the output and input feature maps), is given by:
\begin{equation}
\begin{split}
\label{update_formula}
W^{(t+1)}_{n,c} & = W^{(t)}_{n,c} - \eta\frac{\partial\mathcal{L}}{\partial Y^{(t)}_n}X^{(t)}_c - \frac{\partial\mathcal{R}(W^{(t)}_{n,c})}{\partial W^{(t)}_{n,c}},
\end{split}
\end{equation}

\noindent where $X_c$ and $Y_n$ represent, respectively, the input and output feature maps of the convolutional layer,$\eta$ denotes the learning rate, and $\mathcal{L}$ and $\mathcal{R}$ denote the loss function and weight regularization, respectively. A sparse input feature map, $X_c^t$, may result in a sparse weight, $W_{n,c}^{(t+1)}$, during training. This is because the sparse feature map yields a small weight update The kernel sparsity for the $c^{th}$ input feature map is defined as:
\begin{equation}
    s_c = \sum^{N}_{n=1}\vert W_{n,c}\vert,
\end{equation}
\noindent where $N$ denotes the total number output feature maps.

\subsection{Kernel Entropy ($\mathbf{e_c}$)}
\label{sec:KE}
Kernel entropy is built on the fact that the diversity of the input feature maps is directly related to that of the corresponding convolution kernels. To determine the diversity of the kernels, a nearest neighbor distance matrix, $A_c$, is first computed for the $c^{th}$ convolution kernel. The value in the $i^{th}$ row and $j^{th}$ column of $A_c$ is assigned to be:
\[
    A_{c_{i,j}}= 
\begin{cases}
    \Vert W_{i,c}-W_{j,c}\Vert,& \text{if } W_{j,c}\in \{W_{i,c}\}_k\\
    0,              & \text{otherwise}
\end{cases}
\]
\noindent where $\{W_{i,c}\}_k$ represents the k-nearest-neighbor of $W_{i,c}$. Then, a density metric is calculated for $W_{i,c}$, which is defined as:
\begin{equation}
    dm(W_{i,c}) = \sum^{N}_{j=1}A_{c_{i,j}}.
\end{equation}
\noindent If $dm(W_{i,c})$ is large, then the convolution kernel is more different from the its neighbors, and vice versa. The kernel entropy is calculated as the entropy of the density metrics: 
\begin{equation}
    e_c = -\sum^{N}_{i=1}\frac{dm(W_{i,c})}{\sum_{i=1}^{N}dm(W_{i,c})}\log_2\frac{dm(W_{i,c})}{\sum_{i=1}^{N}dm(W_{i,c})}.
\end{equation}
\noindent A small $e_c$ indicates diverse convolution kernels. Thus, the corresponding input feature map provides more information to the ConvNet.
\subsection{Kernel Sparsity \& Entropy (KSE)}
KSE is then defined as:
\begin{equation}
    KSE = \sqrt{\frac{s_c}{1+\alpha e_c}},
\end{equation}
\noindent where $KSE$, $s_c$, and $e_c$ are normalized into [0, 1], and $\alpha$ is a parameter for controlling weight between $s_c$ and $e_c$, which is set to 1. 

\subsection{Evaluation Metrics}
For quantitative evaluation of the denoised v2 whole-body scans, the ensemble bias in the mean standard uptake value (SUV) of the simulated tumor that was inserted in a real patient background, and the liver coefficient of variation (CoV) were calculated from 10 noise realizations. The ensemble bias is formulated as:
\begin{equation}
    \text{BIAS(\%)}=\frac{\frac{1}{R}\sum_r^R\mu_r^L-T^L}{T^L} \times 100,
\end{equation}
\noindent where $\mu_r^L$ denotes the average counts within the lesion $L$ of the $r^{th}$ noise realization, and $T^L$ represents the "true" (from high quality PET scan) intensity value within the lesion.

The liver CoV was computed as:
\begin{equation}
    \text{CoV(\%)}=\frac{\frac{1}{N}\sum_{i\in B}\sigma^R_j}{\bar{\mu}_B} \times 100,
\end{equation}
\noindent where $\sigma^R_j$ denotes the ensemble standard deviation of $j^{th}$ voxel across $R$ ($R=10$) realizations, $N$ is the total number of voxels in the background volume-of-interest (VOI) $B$. The liver $CoV$ is computed within a hand-drawn 3D VOI within the liver.
\vspace{-1em}
\subsection{Comparisons of The Training Time}
\begin{table}
\vspace{-1em}
\small
\begin{center}
\begin{tabular}{c|c|c|c|c}
\hline
Method & Img. Recon. & Network Training & Total Time & Percent Time Saved\\
\hline\hline
$\text{v1/v2-net}$ & 1 $\frac{\text{wk.}}{\text{Pt.}}$ $\times$ 20 Pts. & 5.5 days & 20.8 wks. & -\\
\hline
\Xhline{3\arrayrulewidth}
$\text{FT/TGD-net}$& 1 $\frac{\text{wk.}}{\text{Pt.}}$ $\times$ 7 Pts. & 2.5 days & 7.4 wks. & 64\%\\
\hline
\end{tabular}
\end{center}
\caption{Comparisons of data preparation and network training time used between the proposed method and training-from-scratch. "FT" denotes "fine-tuning", and "Wk." and "Pt." stand for, respectively, "week" and "patient". To form a complete dataset, it required approximately a week to reconstruct training pairs of noisy inputs and target for each patient (1 target + 6 different count levels), and a total of 20 patients were used for training $\text{v1-net}$ and $\text{v2-net}$.\label{time}}
\end{table}

\newpage
\subsection{Additional Results}
\begin{figure*}[h]
\begin{center}
\includegraphics[width=1\textwidth]{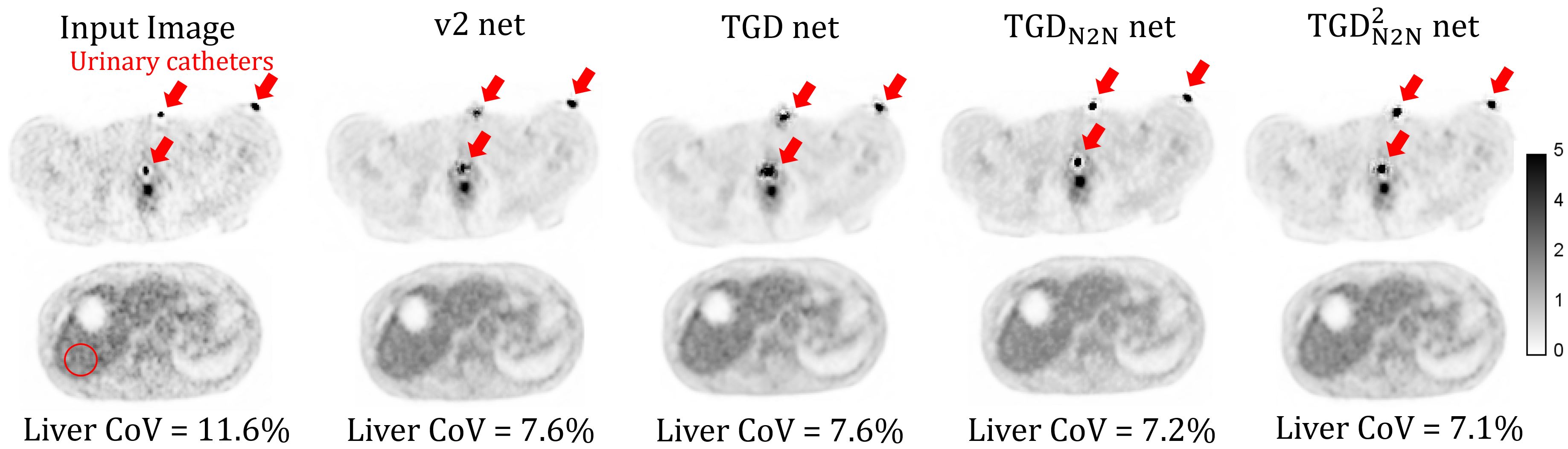}
\end{center}
   \caption{Comparisons of proposed TGD-N2N and other methods on a FDG-PET patient study which had urinary catheters attached during the scan (denoted by the arrows). The first and second rows show the trans-axial slices of the urinary catheters and liver of the same study, respectively. All the images are displayed in the same inverted grey scale. This study was acquired for 120-sec, which was rebinned into 2 noise samples with equal count levels (60-sec) for the TGD-N2N training. Both TGD-N2N networks  were retrained for 150 epoch. All the networks were then applied to the 120-sec scan (input image) to generate the denoised results. The out-of-distribution objects (catheters) led to artifacts in both v2-net and TGD-net results. The online learning approaches using TGD$_\text{N2N}$-net$^{\phi=0.36}$ and TGD$^2_\text{N2N}$-net$^{\phi=0.3,0.4}$ alleviated the artifacts while retaining similar denoising performance in terms of liver Coefficient-of-Variations (CoV) in the ROI denoted by the red circle. The KSE threshold for both TGD-N2N results were adjusted to achieve similar liver CoV for a fair comparison. }
\label{fig:TGD_n2n}
\end{figure*}

\end{document}